\documentclass[manuscript]{acmart}

\AtBeginDocument{%
  }

\setcopyright{acmlicensed}
\copyrightyear{2018}
\acmYear{2018}
\acmDOI{XXXXXXX.XXXXXXX}

\acmConference[Conference acronym 'XX]{Make sure to enter the correct
  conference title from your rights confirmation emai}{June 03--05,
  2018}{Woodstock, NY}
\acmISBN{978-1-4503-XXXX-X/18/06}

\begin{document}

\title{VicSim: Enhancing Victim Simulation with Emotional and Linguistic Fidelity
}


% \author{Yerong Li}
% \affiliation{%
%   \institution{University of Illinois Urbana-Champaign}
%   % \city{Anonymous City}

%   }
% % \country{United States}
% \email{yerong2@illinois.edu}

% \author{Yiren Liu}
% \affiliation{%
%   \institution{University of Illinois Urbana-Champaign}
%   % \city{Anonymous City}
%   % \country{United States}
  
% }
% \email{yirenl2@illinois.edu}

% \author{Yun Huang}
% \affiliation{%
%   \institution{University of Illinois Urbana-Champaign}
%   % \city{Anonymous City}
%   % \country{Anonymous Country}
%   % \country{United States}
% }
% \email{yunhuang@illinois.edu}

% \affiliation{%

% \affiliation{%

\author{Yerong Li}
\affiliation{%
  \institution{University of Illinois Urbana-Champaign}
  \city{Champaign}
  % \state{IL}
  \country{USA}}
\email{yerong2@illinois.edu}

\author{Yiren Liu}
\affiliation{%
  \institution{University of Illinois Urbana-Champaign}
  \city{Champaign}
  \state{IL}
  \country{USA}}
\email{yirenl2@illinois.edu}

\author{Yun Huang}
\affiliation{%
  \institution{University of Illinois Urbana-Champaign}
  \city{Champaign}
  \state{IL}
  \country{USA}}
\email{yunhuang@illinois.edu}

% \affiliation{%

% \affiliation{%


% \newcommand{\todo}[1]{\textcolor{blue}{#1}}
\newcommand{\todo}[1]{\textcolor{black}{#1}}
\begin{abstract}
Scenario-based training has been widely adopted in many public service sectors. Recent advancements in Large Language Models (LLMs) have shown promise in simulating diverse personas to create these training scenarios. However, little is known about how LLMs can be developed to simulate victims for scenario-based training purposes. In this paper, we introduce VicSim (victim simulator), a novel model that addresses three key dimensions of user simulation: informational faithfulness, emotional dynamics, and language style (e.g., grammar usage). We pioneer the integration of scenario-based victim modeling with GAN-based training workflow and key-information-based prompting, aiming to enhance the realism of simulated victims. Our adversarial training approach teaches the discriminator to recognize grammar and emotional cues as reliable indicators of synthetic content. According to evaluations by human raters, the VicSim model outperforms \todo{GPT-4 in terms of human-likeness}.
\end{abstract}



% \keywords{Large Language Models, Scenario-based Training, User Simulation, Safety Reporting, Dialogue System}

\received{20 February 2007}
\received[revised]{12 March 2009}
\received[accepted]{5 June 2009}

\maketitle
\subsection*{}
\begin{figure*}[ht]
    \centering
    \includegraphics[width=0.95\textwidth]{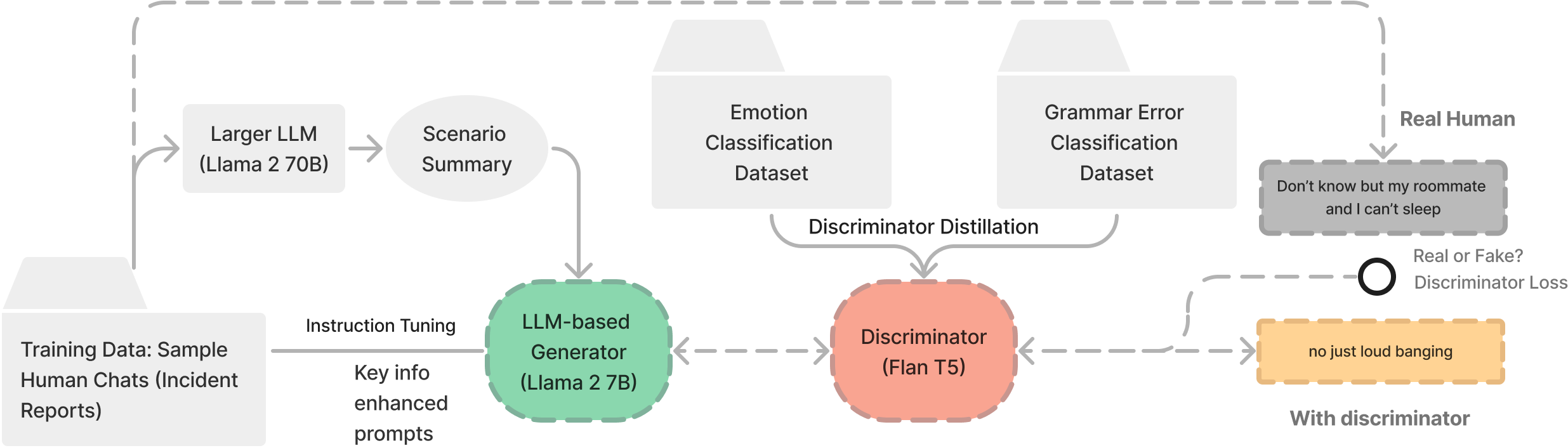}
    \caption{GAN training workflow : we used Flan-T5 based discriminator and Llama-2 chat-based simulated user as the generator}
    \label{fig:GAN}
\end{figure*}

\section{Introduction}
% \yiren{Intro needs more citations}
Scenario-based training has been widely used in many professional domains to enable trainees to acquire complex skills \cite{peeters2014design, salas2006scenario} that are often unable to be learned through traditional lecturing or textbooks. Such training involves asking participants to practice skills by performing tasks designed with learning objectives contextualized in simulated scenarios. 
However, scenario-based training usually requires extensive human resources and efforts from educators in designing and running high-quality scenarios \cite{peeters2014design}. More specifically for public service providers who need to communicate with the general public, such as dispatchers and text-based safety incident reporting \cite{ming2021examining, liu2023discovering}, scenario-based training often involves simulated interactions with service users, typically role-played by training facilitators, to prepare providers for a wide range of real-world scenarios.

Computer-supported training has been widely adopted in many public sectors \cite{reuter2009avoiding, kolmanivc2013sin, reuter2009computer}. 
With recent advances in NLP research, natural language generation methods (e.g., LLMs) have been increasingly used to enrich training scenarios \cite{lin2023using, demasi2020multi}.
In the domain of AI/ML, user simulation aims to extract valuable insights and patterns from user behaviors in different in-context scenarios~\cite{10.1145/3624918.3629549,tan2023user}. 
This process is instrumental in adapting systems to cater to the unique requirements of specific user groups.

This aspect of user modeling is particularly critical but challenging in the context of safety incident reporting, given victims tend to demonstrate negative emotion fluctuations while experiencing safety-related incidents and exhibit varied reporting behavior~\cite{feldman2021calming, tracy1998emotion, liu2023discovering}. 
In this context, user modeling can help enhance the customization and adaptation of training systems and improve dispatchers' preparation for handling real-life incident reports with both accuracy and empathy. 
% User modeling techniques can be applied to extract their patterns.
%Leveraging simulated interactions has the potential to train dispatchers in realistic scenarios to acquire and improve crucial skills and learn from diverse simulated scenarios.
However, to the best of our knowledge, there is no LLM-based simulator system for training dispatchers who handle incident reports. 
% Thus, the accuracy and effectiveness of report handling along with the management of emotional interactions can have significant implications~\cite{veale2023no,ask2010emotions}.

\begin{comment}
User modeling techniques have facilitated a better understanding of user behaviors, and customized intelligent assistance, and can greatly improve user experience.

In text-based safety reporting systems, where users may seek emotional support in distressing situations, the role of user simulation becomes increasingly critical. Effective scenario-based user models can aid in training dispatchers, teaching them how to deliver emotional support and gather key information during interactions. Simulation-based training could be essential to emergency workers as dispatchers, who are regularly exposed to traumatic events with potential physical and emotional burnouts~\cite{golding2017exploring,klimley2018posttraumatic, smith2019exploring, lawn2020effects}. 
Despite their developed tolerance from working experience, they still struggle with the effects of trauma because of the demanding nature of their job and limited knowledge in every individual emergency case. The necessity to comprehend the nature and impact of traumatic incidents, as emphasized by recent literature \cite{tsirimokou2024experiences, cramm2021mental}, underscores the importance of user modeling in facilitating effective communication and support provision in text-based safety reporting systems.
\end{comment}

In this paper, we focus on simulating "text-based incident" reporting. We introduced several enhancements to the scenario-based user simulation in the safety incident reporting domain. Building upon Llama 2-7B-based chat model~\cite{touvron2023llama}, we have developed a suite of improvements aimed at elevating both the informational and emotional faithfulness of simulated users. 
Our methodology involves a multifaceted approach to guide dialogue generation to enhance the authenticity of user responses. With these improvements, our objective is to develop a dialogue experience that is more detailed and engaging, fostering deeper interactions within scenario-based applications. Specifically, by leveraging these enhancements, we seek to simulate a more human-like user based on constructed safety reporting incidents. This enables the chatbot to serve as a valuable tool for dispatcher training, allowing dispatcher trainees to engage in conversations with a simulated user that closely mimics real-world scenarios. 
Our contribution includes the new model and empirical understanding of the effectiveness of the model in the potential scenario-based training context. 

Specifically, we address the following research questions:

% \yun{I will revisit the RQs after skimming through the whole paper.}
\begin{itemize}
    \item \textbf{RQ1}: \textit{How does scenario-based LLM user simulation differ from humans in terms of chat style and informational faithfulness in safety incident reporting?}
    % \item \textbf{RQ2}: \textit{When do users ask for emotional supports?}
    \item \textbf{RQ2}: \textit{In what ways can we improve the emotional dynamics of the victim simulation model?}
    \item \textbf{RQ3}: \textit{In what ways can we better simulate human-like victims in incident reporting interactions in terms of chat style?}
\end{itemize}

This research has practical implications that the contributed model \textit{VicSim} can be applied to enrich scenario-based training for dispatchers in diverse settings to acquire and improve crucial skills and learn from diverse simulated scenarios.
\section{Related Works}

\subsection{User Simulation}

User simulation plays a crucial role in the understanding of user behaviors, facilitating the customization and adaptation of response systems tailored to a group of prototype users. It enables us to gain deeper insights into user behavior, allowing for the refinement of system responses. Additionally, user simulation aids in training system staff to effectively respond to potential users across various scenarios.

Through user simulation techniques, researchers have gained a deeper understanding of user behaviors, leading to customized intelligent assistance and significantly enhanced user experiences. For instance, when individuals search for an online course through customer service, user simulation techniques analyze their search history, predict their current learning interests, and provide personalized recommendations tailored to their preferences and learning goals. These simulations take into account factors such as past course selections, browsing patterns, engagement levels, course selection from users with similar query histories, and feedback to deliver relevant and engaging course suggestions. By leveraging user simulation, platforms can better cater to the diverse needs and interests of learners, ultimately improving the overall online learning experience. The impact of user simulation extends across various domains such as e-commerce~\cite{berger2019user, vullam2023multi, li2022user, deldjoo2024fairness}, social networks~\cite{gatti2014large,birdsey2015twitter,li2017generating}, search engine~\cite{sallesearch2021}, task-oriented dialogues~\cite{li2016user, gur2018user,terragni2023context}, education~\cite{de2013simulation}, etc. making it a highly active and influential research field.

User simulation typically involves mining and learning from user data, which includes user profiles, user-generated content, and user interactions with content and other users. User profiles are often characterized by attributed interests or personas. User-generated content encompasses diverse text data sources such as conversation histories, reviews, blogs, etc. which can be analyzed using natural language processing techniques. User interactions include actions as rating and sharing, which represent edges between nodes, e.g., users, in text-attributed graphs. Graph mining techniques are accordingly employed to analyze such interactions.~\cite{beutel2016user,wang2019exploring} Consequently, user simulation leverages a combination of text-based and graph-based approaches, focusing on mimicking real humans from both text and graph data.

The advent of transformers has revolutionized user simulation research, particularly with pretrained language models operating under the pretrain-finetune paradigm. Pretrained language models leverage self-supervised learning on large unlabeled corpora and fine-tuning on specific tasks with a smaller labeled dataset. These models have showcased remarkable abilities in reasoning, generalization, and knowledge comprehension, surpassing human performance in various tasks. Their emergence has sparked a new era in User Simulation research, prompting a reevaluation of user-generated content mining strategies.

\subsection{AI-based Solutions for Simulating Scenario-based Interactions}
AI-based systems have been widely adopted to facilitate the delivery of services in public sectors, such as child protection \cite{vaithianathan2017developing}, homeless services \cite{kuo2023understanding}, and public employment \cite{scott2022algorithmic}.
Recently, studies have also explored the use of LLMs in various public service domains, such as public health \cite{jo2023understanding} and digital government \cite{gao2023application}. 

Most service providers require extensive training for employees. Recent developments in AI/ML algorithms have enabled application in training professional practitioners of varied domains. For instance, in the medical domain, practitioners have adopted ML-based technology for medical professional training in robotic surgery simulations \cite{yamashita2016remote, lomis2021artificial} and ECG reading \cite{tahri2021interpretation}. 
The repetitive nature of such training processes improves participants' learning outcomes and fosters the robust ability to adapt in context-specific situations~\cite{lomis2021artificial}. 
However, facilitating repetitions in scenario-based training usually incurs high costs in terms of experienced staff members required during training~\cite{peeters2014design}. Additionally, for scenarios that require social interactions with other humans or service users, recruiting role-players or actors that have the capability to simulate real users is even more challenging and resource-consuming. Scenario actors are often required to possess the ability to adapt to changes during the scenario simulation, and react to trainees' actions while sticking to the learning objectives.

In the context of victims reporting incidents, the delivery of emotional support by dispatchers is crucial, especially considering the emotional distress individuals may experience when reporting traumatic or safety-related incidents \cite{Observations_on_the_Display}. Proficient emotional support not only aids in information gathering but also enhances trust and cooperation from callers. On the other hand, inadequate emotional support can strain the dispatcher-caller relationship, potentially leading to user frustration and reluctance to cooperate \cite{paoletti2012operators,feldman2021calming}. This underscores the importance of training dispatchers to adapt to users' emotions in different scenarios. Simulating users' emotions in various scenarios is thus vital for dispatcher training, ensuring they can effectively respond to callers' emotional needs while gathering necessary information for incident handling. While existing research has primarily focused on the functional aspects of text-based reporting systems, there is a gap in understanding how dispatchers should navigate and address users' emotional states within such systems to optimize service delivery. This highlights the need for further empirical research and training initiatives aimed at enhancing dispatcher capabilities in emotional support in the safety incident reporting domain. 

Recent research on LLMs has demonstrated their potential in simulating users and personas \cite{jiang2023personallm, salewski2024context}. Yet, little research has been done on how LLMs could be leveraged for simulating victims sending incident reports, such that the simulated messages can be used for scenario-based training.

\section{Method}

\subsection{\textit{VicSim}: 
Simulating Victims' Text Reporting} 
Figure \ref{fig:GAN} illustrates the data processing and training workflow of \textit{VicSim}. We observe distinctions between the LLM-based agent and human beings in terms of emotional dynamics and grammar styles. To address this, we fine-tuned the FLAN-T5 based discriminator on two datasets—one for emotional classification and one for grammar classification. This approach allows us to obtain a more refined text representation with a heightened focus on grammatical and emotional aspects of sentences. With this refined discriminator, we aim to improve the identification of fake and real sentences, thereby guiding a better LLM-based generator.

To initiate the fine-tuning of the generator, we constructed a prompt illustrated in Figure \ref{fig:alignment}. The prompt begins with system messages, followed by scenario summaries generated by the Llama-2 70B model outlining a user’s case. Dialogue history is then presented after the scenario, as demonstrated in the example figure. This combined prompt directs the LLM to generate simulated user responses tailored to the given scenario, incorporating both dialogue history context and summarized user experience.

In order to simulate victims who sent text-based reports, we fine-tuned the Llama-2 7B chat model by aligning the chat history with the user's utterances. Specifically, we start by prompting the LLM with both system prompt on how to simulate a user and the scenario summaries on a user's case. The prompt begins with system messages, followed by scenario summaries generated by the Llama-2 70B model outlining a user's case. Dialogue history comes after the scenario, as shown in Figure \ref{fig:alignment}. This combined prompt directs the LLM to generate simulated user responses tailored to the given scenario with both dialogue history context and summarized user experience.

\begin{figure*}
\small
\resizebox{\textwidth}{!}{%
\begin{tabular}{ p{.2\textwidth} p{.6\textwidth}}
\hline
 System Prompt &
Imagine you're in a situation where you need to report a safety concern to your local police department. You're speaking with a dispatcher from the police department, who is trained to provide helpful and detailed guidance, as well as necessary emotional support. Describe your interaction with the dispatcher, including what safety concern you're reporting and how the dispatcher assists you.\\ Scenario &
\textbf{The user reported seeing a student riding their stolen bike on campus. The user provided a detailed description of the suspect, including their physical appearance (white male, dark brown/black hair, jeans and a tan-colored pullover), and the time and location of the incident (about 2 hours ago, on campus).}\\\\
Dialogue History &
\parbox[t]{.6\textwidth}{User:  I saw a student riding my stolen bike on campus. He was riding toward [John Smith]. My bike is pink and green with a basket on it. 
 \\Dispatcher:  How long ago did you see them riding your bike? 
 \\User:  I should have taken a picture but I just froze when I saw my stolen bike. It was stolen right before Christmas. 
 \\User:  About 2 hours ago. I had to leave to go to a meeting on the hill 
 \\Dispatcher:  Can you describe the person you saw on your bike? 
 \\......\\
} 
\\ \hline
 User Response & {User: White male, dark brown/black hair that was almost shoulder length. He was wearing jeans and a tan-colored pullover}\\ \hline
\end{tabular}}
\caption{Illustration of the prompt construction process for simulated user dialogue generation: we concatenate the system guidance, user's scenario, along with dialogue history, enabling comprehensive prompts for dialogue generation by the LLM}
\label{fig:alignment}
\end{figure*}

% \subsection{Techniques to improve Chat Style Consistency}
% \begin{figure}[ht]
%     \centering
%     \includegraphics[width=0.5\textwidth]{figures/GAN.drawio.png}
%     \caption{GAN training workflow : we used Flan-T5 encoder based discriminator and Llama-2 chat-based simulated user as the generator}
%     \label{fig:GAN}
% \end{figure}

To encourage human-like generations, we adapted an adversarial training approach by retraining the Llama-2-based user as a generator in a generative adversarial network (GAN)~\cite{NIPS2014gan}, as shown in Figure \ref{fig:GAN}. This involves leveraging a Flan-T5-based classifier \cite{longpre2023flan} as the discriminator and the Llama-2-based user as the generator. 

To improve the text representation of the Flan-T5-based encoder, we instruction-tuned Flan-T5 with emotional prompts and prompts designed to simulate grammar errors. These prompts were used as instructions to align with the emotional and grammar error labels with the dataset described in section \ref{sc:evaluation}. By incorporating emotional prompts, we aimed to imbue the encoder with a deeper understanding of nuanced emotional expressions within the text. Additionally, by simulating grammar errors, we sought to train an encoder that can recognize and potentially correct grammatical inconsistencies, thereby improving overall text coherence and quality. Through this meticulous tuning process, we aimed to refine the Flan-T5-based encoder to better capture relevant representations of text data.

Specifically, given a context prompt $p$, $G(p)$ represents the output of the generator network, a.k.a. the simulated user. It is expected to produce synthetic data samples, i.e., fake sentences. In discriminator loss $L^D_{\text{GAN}}$, the first component, represents the expected value of the negative logarithm of the discriminator's output when given real data samples $x$. This term encourages the discriminator to assign high probabilities to real data samples. While the second component, 
$\mathbb{E}[\log(1 - D(G(p)))]$, represents the expected value of the negative logarithm of the discriminator's output when given generated data samples 
$G(p)$. This term encourages the discriminator to assign low probabilities to utterances generated by LLM. The generator aims to minimize this term by generating samples that the discriminator is more likely to classify as real.

\begin{equation}
L_{\text{GAN}}^G = -\mathbb{E}[\log D(G(p))]    
\end{equation}

\begin{equation}
L_{\text{GAN}}^D = -\mathbb{E}[\log D(x)] - \mathbb{E}[\log(1 - D(G(p)))]
\end{equation}
% One of the key challenges we encountered was the imbalanced learning rates between the generator and the discriminator. To address this, we implemented a strategy where the generator's learning rate was set to be ten times higher than that of the T5-based discriminator. This adjustment helped in mitigating the discrepancies in learning speeds, ultimately contributing to a more harmonized training flow between the discriminator and the generator. Grammatical errors are common in human's utterances. As we can see in a later analysis, a human user rarely speaks an error-free sentence, with fewer than 6\% of the user utterances containing grammatical mistakes. 
% \begin{quote}
% \begin{tabular}{|l|}
% \hline
% \\
%     \parbox[t]{.4\textwidth}{... You may be feeling nervous, which can lead to grammar errors such as punctuation errors, spelling mistakes, and sentence structure errors etc.}\\\\
% \hline
% \end{tabular}
% \end{quote}

%\subsection{Improving Information Faithfulness} \yun{this section is too tiny. you could make it a subsubsection, but it doesn't seem providng enough details on what you did. }
To reduce hallucination, we augmented a prompt with key information extracted from the CoreNLP toolkit. This way, we aim to alleviate hallucination and encourage a more focused user response. Figure \ref{fig:cocatner} shows how we listed the attributed keywords with a given scenario. Detailed analysis of this augmentation is in Section \ref{sec:infofaith}.
\begin{figure}
\resizebox{.4\textwidth}{!}{%
\begin{tabular}{p{.1\textwidth} p{.3\textwidth}}
\hline
Scenario           & The user was concerned about loud banging noises coming from upstairs, which they believed to be in the second floor of their building. The user provided their location as "courtyard Apartments" and shared details about the noise, including that it was coming from above them and that it happened during nighttime and afternoon. The user also mentioned that they were a student at UCLA and that they had not reported the noise to the university. The dispatcher sent an officer to the location and the user spoke with the officer, Jane. The officer confirmed that the noise was coming from upstairs and that they would investigate further.
\\\hline
&ORDINAL : second

TIME : nighttime

TIME : afternoon

TITLE : student

ORGANIZATION : UCLA

PERSON : Jane
\\\hline

\end{tabular}}
\caption{Senario-based prompt argumented by the key information extracted from the CoreNLP toolkit}
\label{fig:cocatner}
\end{figure}

% \subsection{Language Style?}

\subsection{Dataset }
\subsubsection{Data Collection}
\begin{figure}
    \centering
    \includegraphics[width=0.7\textwidth]{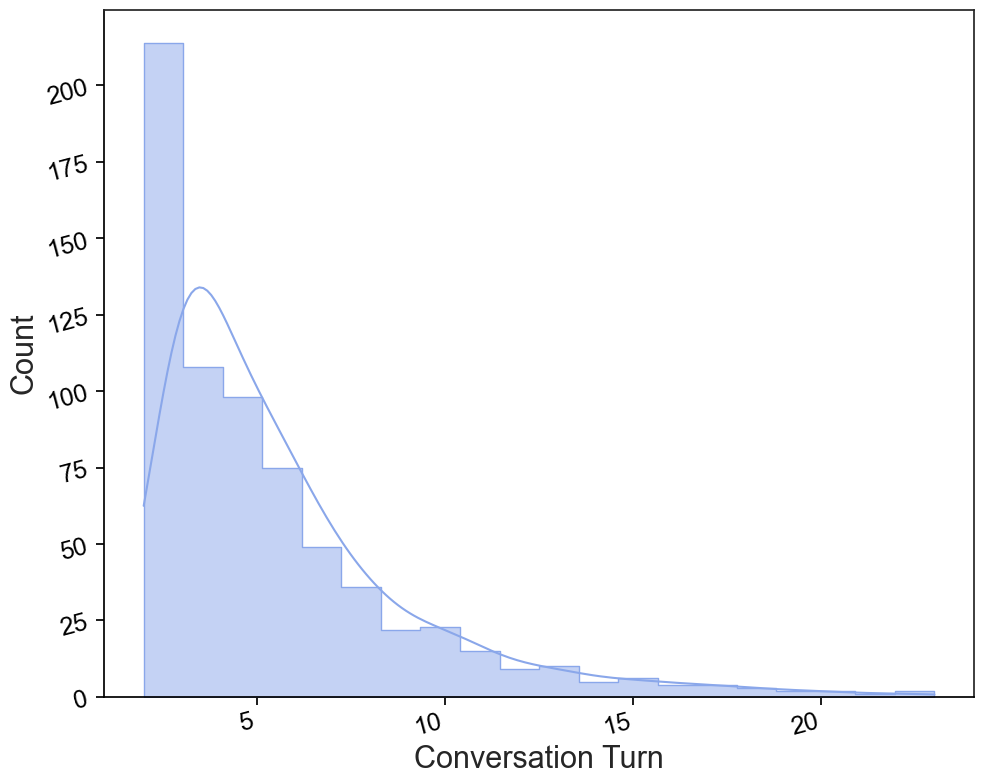}
    \caption{Users' utterances at different conversation stages on the evaluation set}
    \label{fig:eval-length}
\end{figure}

The dataset used in this study comprises chat logs collected through a text-based incident reporting system, which has been adopted by over 200 higher education institutions~\cite{ming2021examining}. The system facilitates communication between members of an organization and safety teams responsible for risk management \cite{ming2021examining}.  Community members can submit information or "tips" via their mobile app or web portal. Safety organization dispatchers, in turn, can respond to these tips using the dashboard. Tips may include text, photographs, video, or audio recordings and can be categorized into different "tip types," selected by local safety departments for each organization. We eliminated "Test" entries and retained only relevant categories. Specifically, we preserved nine well-populated, detail-rich event types, such as SuspiciousActivity, DrugsAlcohol, and NoiseDisturbance, etc. We retained chat logs committed between January 1, 2018, and December 31, 2019, and retained chats with more than two utterances (or one conversation turn), as we focused on investigating chats involving at least one turn of information-collecting conversation. We then evaluated users' utterances in different conversation stages, from turn 2 to turn 22, as illustrated in Figure \ref{fig:eval-length}.

\subsubsection{Data Ethics}
% \todo{TODO}
The study was carried out under a strict non-disclosure agreement (NDA), analyzing anonymized data from consenting users and organizations for research on system improvements. The data was thoroughly anonymized, removing or masking all personal and sensitive information, including identities, geolocations, and organizational details, with anonymous IDs and generic tags (e.g., [LOCATION], [NAME]). The masked information was then synthesized and filled in using Llama 2-70b. This approach significantly reduced de-anonymization risks.
Additionally, as a further privacy measure, data from a random selection of the institutions was excluded from the dataset prior to researchers' access.

\begin{table}[ht]

\begin{tabular}{lll}
\hline
Event Type             & \# of User Utter. & Avg. length\\\hline
SuspiciousActivity     & 1227       & 6.98      \\
AccidentTrafficParking & 116        & 6.24      \\
DrugsAlcohol           & 618        & 6.23      \\
EmergencyMessage       & 1032       & 7.06      \\
FacilitiesMaintenance  & 157        & 5.26      \\
HarassmentAbuse        & 431        & 7.19      \\
MentalHealth           & 242        & 8.00      \\
NoiseDisturbance       & 1359       & 6.12      \\
TheftLostItem          & 245        & 7.86     \\ \hline
TOTAL                  & 5427       & 6.48     \\ \hline
\end{tabular}
\caption{Statistics on the number of user utterances across 9 different incident types in this study}
\end{table}

\subsection{Evaluation}
\label{sc:evaluation}

We evaluated \textit{VicSim} from three key areas: user emotion understanding, grammar style fidelity, and context faithfulness. Additionally, we complement these three automatic evaluations with human judgment to gain deeper insights into the subjective aspects of response quality. Human evaluation allows us to assess nuances such as fluency, coherence, and overall naturalness, which are essential for generating engaging and contextually relevant conversations.
% Nevertheless, for incidents characterized by more complex issues and a greater need for human empathy \cite{zhao2023chatgpt}, such as \textit{Mental Health} incidents,  human dispatchers may refer less to AI models.  
% Our find

% \yiren{Add a distribution of the depth of utterances that are being evaluated}

\subsubsection{Assessment of Informational faithfulness} 
In the assessment of context faithfulness, we employed NER modules within Stanford CoreNLP\cite{manning-etal-2014-stanford} as a proxy to extract keywords from the scenarios generated by the Llama-2 70B. Subsequently, as depicted in Figure \ref{fig:assess_hallu} we calculated the overlap between these extracted keywords and those present in a user's utterances. A successful recall is identified when a user mentioned a keyword that appeared in groundtruth keywords in the scenario summary. This approach allowed us to gauge the fidelity of the generated scenarios in capturing relevant contextual information, as indicated by the presence of matching keywords in user interactions. As shown in the example demonstrated in Figure \ref{fig:assess_hallu}, the utterance generated by LLM successfully retrieved the ground truth keywords 'Cortright' and 'teacher' in the outlined scenario.

\subsubsection{Assessment of User Emotions} 
We fine-tuned a RoBERTa-based classifier on the GoEmotions dataset, which classifies emotions into 28 fine-grained emotions. We follow the positive / negative / neutral emotion classification as described in the GoEmotions dataset. 
% With a 20\% training and validation split, our BERT-based classifier achieves a 46.34\% classification accuracy on the validation set.
We evaluate the emotional faithfulness of the user simulation model based on the emotional dynamics within the generated responses. Given a scenario with ground truth conversations, we compare the classified emotion categories between each model-generated response with the ground truth. 
% Whenever a user utterance is associated with negative emotions classified by RoBERTa, a need for emotional support arises. LLM or the human successfully provides emotional support whenever it generates a positive response to a user’s negative utterance.
\subsubsection{Assessment of Grammar styles} 
% \yun{I feel some of these are method, not the evaluation part?}
Humans make grammar mistakes, while language models don't. To gauge this difference, We fine-tuned the RoBERTa-based model on a grammar classification error dataset\cite{KaggleGrammarCorrection} along 2,000 grammatically correct sentences from C4\_200M\cite{stahlberg-kumar-2021-synthetic}. The dataset encompasses 37 categories, which include 36 types of common grammar errors as well as a category for sentences with no errors. Through fine-tuning, we constructed a three-layer linear classification head on top of the RoBERTa output. This proxy evaluation model achieved an accuracy of 94.25\% on the partitioned validation set.

\begin{figure}
\resizebox{.4\textwidth}{!}{%
\begin{tabular}{p{.1\textwidth} p{.3\textwidth}}
\hline
Scenario           & The user reported a classmate who was extremely disrespectful towards a teacher and made the entire class super uncomfortable by pacing the classroom and crying and screaming and shouting. The user provided the classroom number ([Dante] room \#\#\#) and the professor's name [Cortright].\\
keywords&\{'Dante', 'Cortright', 'professor', 'teacher'\}\\\\\hline Context& 
\parbox[t]{.3\textwidth}{User:  I saw a student riding my stolen bike on campus. He was riding toward [John Smith]. My bike is pink and green with a basket on it. 
 \\Dispatcher:  How long ago did you see them riding your bike? 
 \\User:  I should have taken a picture but I just froze when I saw my stolen bike. It was stolen right before Christmas. 
 \\User:  About 2 hours ago. I had to leave to go to a meeting on the hill 
 \\Dispatcher:  Can you describe the person you saw on your bike? 
} 
\\ \hline
Utterance & {User: It was yesterday during our 2nd period class. The teacher is Mrs. Cortright.}\\ keywords & \{'2nd', Cortright, 'yesterday', 'teacher'\}\\ \hline
\end{tabular}}
\caption{Keyword-based assessment on hallucination, here [John Smith], [Cortright] are simulated key information we filled in for data processing.}
\label{fig:assess_hallu}
\end{figure}

\subsubsection{Human Evaluation}

To understand the extent to which our model can simulate the responses of real victims, we created a survey and asked Amazon Mechanical Turk annotators to rate key aspects of the outputs generated by our model in terms of human-likeness.
For each rater, we randomly sampled one incident and presented the dialogue history to the rater.
We then present the last user response generated by our model, human, and GPT-4 to the user respectively, and collect rater's ratings regarding four dimensions including coherency, consistency, level of detail, human likeness. 
We designed a rating survey using 5-point Likert scale items for each dimension.
The responses for each incident are presented in a random order. 
\section{Findings}

\subsection{Informational Faithfulness and Human Likeness of \textit{VicSim} (RQ1)} 

First, we evaluated the informational faithfulness of different models using information extraction results, by comparing the key information present in the generated response with those in the provided scenario.

% \yun{maybe this can be merged with the human raters' results and move up as the first result?}\yiren{Merged and Moved}
\label{sec:infofaith}
\begin{table}[h!]

\begin{tabular}{ccccc}
\hline
   & Human & GPT4 & VicSim w/o keywords & VicSim\\ \hline
P  & 39.39 & 40.24  & 37.42  & 36.74           \\
R  & 31.26 & 13.12  & 19.77  & 22.69           \\
F1 & 34.86 & 19.79  & \textbf{25.87}  & \textbf{28.05}           \\ \hline
\end{tabular}
\caption{Comparsion of human, VicSim and GPT4in terms of delivering key information in the scenario}
\label{tb:context}

\end{table}

\paragraph{\textbf{VicSim surpasses GPT-4 in informational Faithfulness.}}
Table \ref{tb:context} indicated that LLM agents were poorer compared to human users in terms of contextual consistency. Prompting VicSim with keywords helps alleviate hallucination and largely boosts recall rates. As a peer comparison to GPT-4, our locally aligned VicSim, with or without keyword prompting, can more accurately deliver key information such as location and time. GPT-4 suffered from hallucinations in a scenario-based dialogue simulation.

\begin{figure}[ht] 
    \centering
    \includegraphics[width=0.8\textwidth]{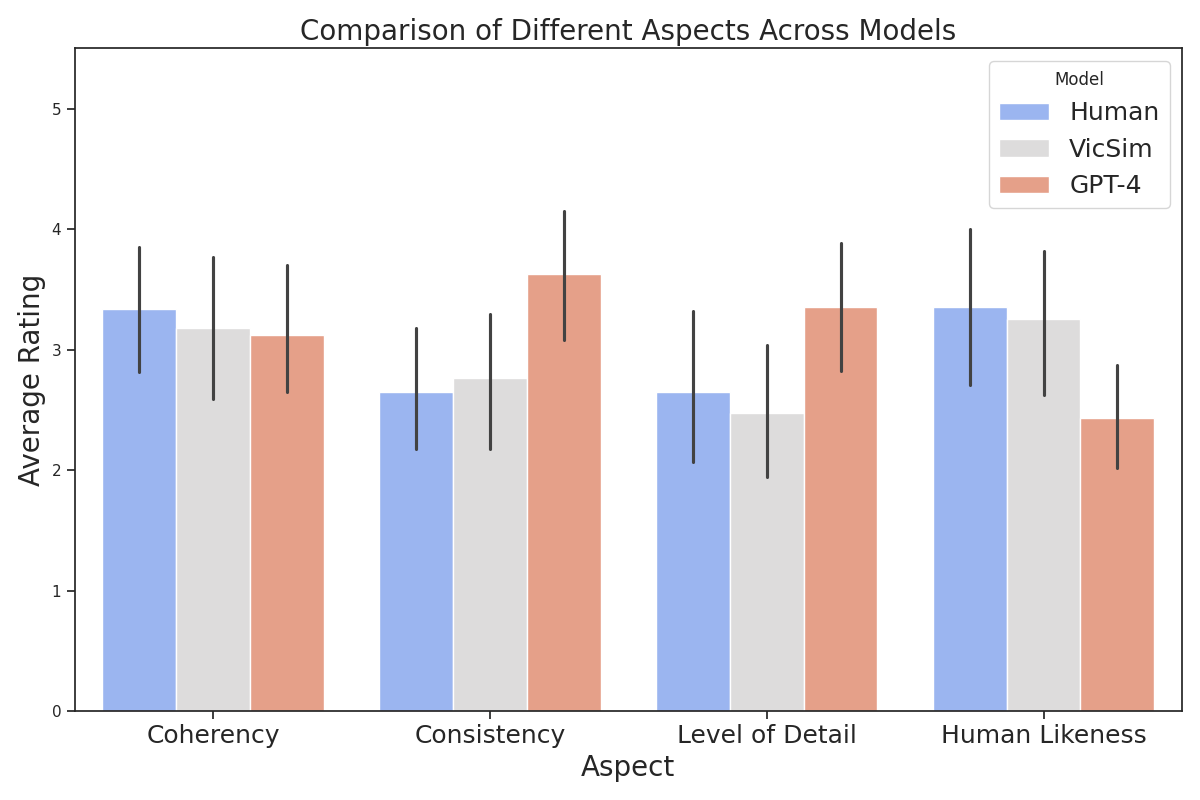}
\caption{Distribution of Ratings from Human Raters; paired t-test indicated a significant difference between the responses from human and GPT-4 ($F=-2.22$,  $p<0.05^{*}$), while no significant difference was found between human and our model.}
    \label{fig:dist-ratings}
\end{figure}

% \subsubsection{Human Rater Evaluation}
\textit{\textbf{Human Evaluation showed no significant difference between VicSim-generated and human-written responses.}}
We analyzed the rating results of model-generated victim responses provided by human raters. The resulting distribution of the ratings for each aspect is shown in Figure \ref{fig:dist-ratings}. VicSim, along with humans, might have lower consistency in conversations compared to GPT-4, but VicSim looks more like human beings. The human evaluation results showed that while the responses generated by VicSim did not exhibit a significant difference from ground truth human written responses in terms of both human and machine likeness, human evaluators rated responses generated by GPT-4 significantly more likely to be generated by AI comparing to human written responses ($F=-2.22$,  $p<0.05^{*}$). This indicated that our proposed VicSim model has the capability to simulate more human-like victim responses than GPT-4. 
By examining the samples rated by human raters, we found that responses with more details that are factually coherent tend to have higher ratings in terms of human likeness. 
For instance, during the report of a Drugs/Alcohol incident, when asked by the dispatcher about the item carried by the offender, the VicSim model generated the response ``\textit{Yes, it was a small, silver metal tube with a blue light on the front. I'm pretty sure it was a crack pipe,}'' while the human wrote ``\textit{It was pipe like. I'm not exactly sure what it is called, but I could clearly tell what it was used for. Maybe a hookah I think?}'' The response provided by VicSim provided more details on the item, and subsequently received a higher human likeness rating than the human written sample. Meanwhile, GPT-4 generated the response ``\textit{It was a small black bag with red stripes on it. It looked like a glass pipe and some small plastic bags inside,}'' which involves commonsense errors, thus leading to the lowest human likeness rating.

Participants also shared their perception of why discerning human responses was challenging. 
Two major difficulties were identified: First, they were unfamiliar with the linguistic patterns exhibited by individuals in real life during urgent or emotionally charged situations; 
Second, while they anticipated that the verbose or detailed nature of responses typically generated by LLMs might serve as a reliable indicator, they were surprised to find that human responses could also be similarly detailed.
Moreover, those participants who were successful in differentiating between human and LLM-generated responses often relied on subtle grammatical cues or specific uses of punctuation.

\paragraph{\textbf{When a scenario summary lacks key information, LLM hallucinates}} When a scenario summary lacks sufficient detailed information, LLM compensates by providing unmentioned key information such as location and the suspects' skin color. Key information may include person-related content, location, time, organization, incident type, etc, increase the likelihood of hallucinating. For example, in the following dialogue, for example, when the dispatcher asks about the missing dogs' appearance, LLM finds no detailed information in the given background and thus hallucinates, providing a description of "black and brown, medium size, and seem to be a mix of German Shepherd and Labrador."  

\begin{quote}
\textbf{Tip category}: NoiseDisturbance

\textbf{System} : ... Scenario: The user reported two dogs that were roaming without their owner in Downtown and Los Angeles. The user provided a detailed description of the dogs, including their colors and breeds. The user also shared that they had posted about the dogs on a Nextdoor page.

\textbf{User:}  Two dogs are roaming without their owner on Main Street and John Smith Drive. around 10:30am. They tried to attack my leashed dog. 

\textbf{Dispatcher:}  I will have an Officer check the area. Can you tell us what they looked like? 
 \begin{itemize}
    \item \textbf{Groundruth keywords} \{'Angeles', 'Los', 'Downtown', 'owner', 'two', 'Nextdoor'\}
    \item \textbf{VicSim:} They are both black and brown, medium size, and seem to be a mix of German Shepherd and Labrador. 
    
    \textbf{keywords} \{'Labrador', 'Shepherd', 'German'\}
    \item \textbf{Human:} One was medium black and white and the other was brown looked like a pit bull bulldog mix.

    \textbf{keywords} \{'One'\}
\end{itemize}
\end{quote}

\subsection{Emotional Dynamics of \textit{VicSim} (RQ2)}
\label{Sec:emotionalDifference}
\begin{figure}[ht]
    \centering
    \includegraphics[width=0.8\textwidth]{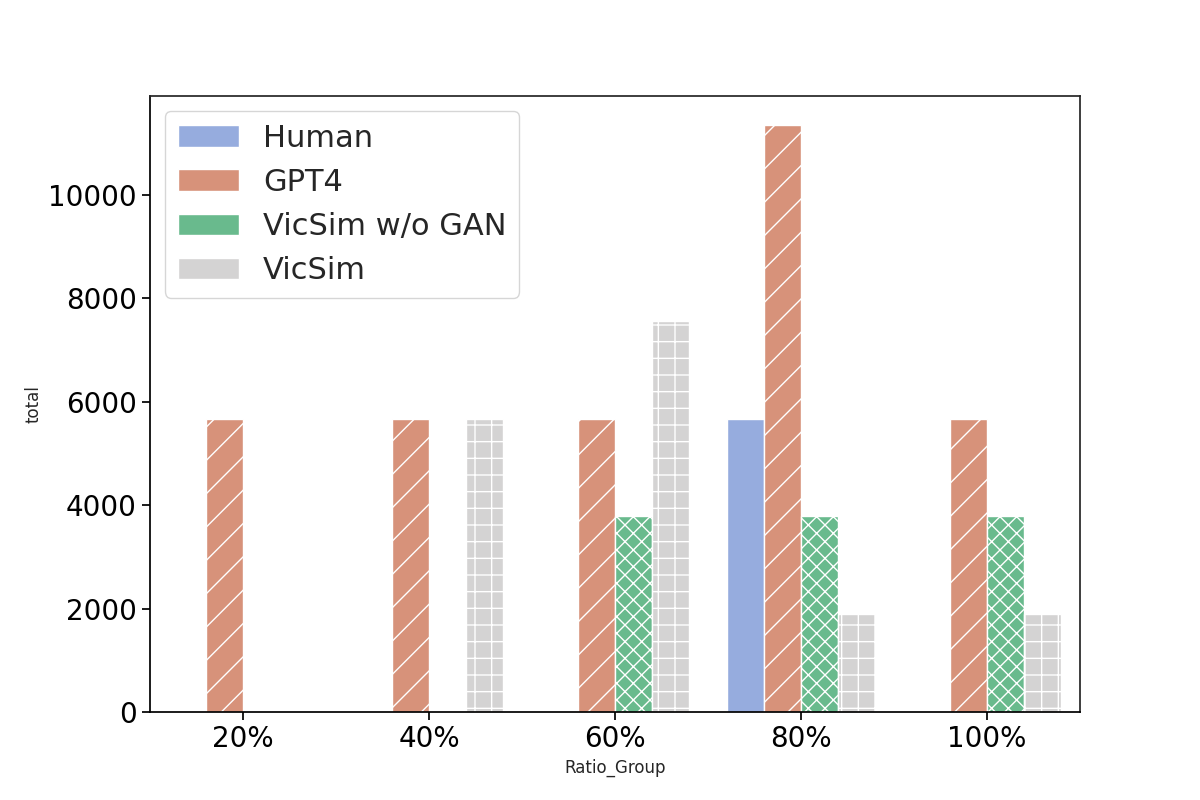}
    \caption{Emotional Trajectories of Human and Simulated Users Throughout Dialogues: with or without adversarial training, local LLM exhibit similar patterns of emotional exposure over time, with a tendency to express more negative feelings during the initial 60\% of the dialogue. By comparison, GPT-4 does not follow human-like evolving patterns in emotions.}
    \label{fig:progress}
\end{figure}

In this section, we present our findings in the emotional dynamics generated by \textit{VicSim} from the following perspectives. First, we presented the different emotional trajectories over the reporting stages between human chats and \textit{VicSim}-generated dialogue. We found that \textit{VicSim} expressed more emotions than humans in general, and then we shared three observations that may explain the differences. For example, we found an association between emotional expressions and the chat length, LLM were more emotional when repeatedly describing situations, and LLM expressed emotions when they hallucinated. We further provide sample dialogues to illustrate our observations about when emotions are expressed by \textit{VicSim}. 

\subsubsection{Different Emotional Expressions between \textit{VicSim} and Humans}
First, regarding emotional changes at different stages of reporting dialogue, with our RoBERTa-based emotion classifier, our analysis reveals a notable pattern in user emotions during safety incident reporting dialogues: users tend to express negative feelings at the beginning of a dialogue, typically when the conversation has progressed to around the 40\% mark of the conversation. GPT-4, however, does not exhibit similar patterns. This finding underscores the significance of emotional support provision at the initial stages of user interaction, suggesting that users often require immediate emotional assistance or reassurance when initiating a dialogue regarding safety incidents. Furthermore, our aligned LLM, mimicking scenario-based user behavior, demonstrates a similar trend, indicating the model's ability to capture and replicate key emotional dynamics observed in real user interactions. This alignment highlights the potential of utilizing AI-driven dialogue systems to effectively respond to users' emotional needs in safety incident reporting scenarios, particularly by prioritizing emotional support delivery during the early stages of interaction. By emulating users' emotional trajectories, these AI models can better serve as training systems to train human dispatchers on how to respond in scenarios that involve evolving emotions. As shown in Figure \ref{fig:progress}, LLM generated more emotions compared to the human, we have several insights on this.
\begin{figure}[ht]
    \centering
    \includegraphics[width=0.6\textwidth]{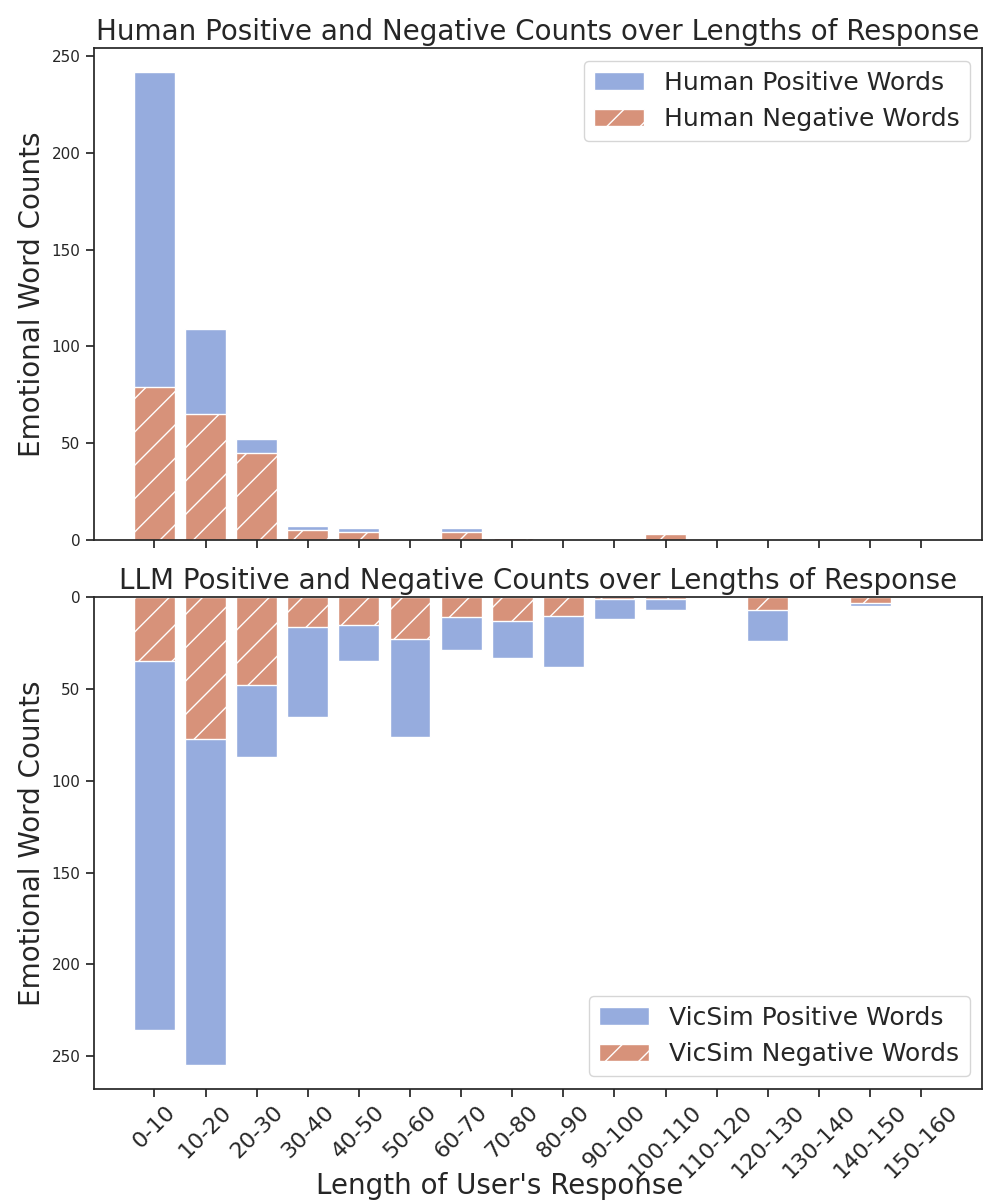}
\caption{Difference in distribution of emotional keywords over human/LLM responses of different lengths }
    \label{fig:dist-length}
\end{figure}

\subsubsection{Three Observations of \textit{VicSim}'s Emotional Expressions}
\paragraph{\textbf{LLM generates longer responses: the longer, the more emotional}}
% \begin{table}[]
% \begin{tabular}{|l|l|l|}
% \hline
% \begin{tabular}[c]{@{}l@{}}Pearson Cor.\\ sentence length\end{tabular} & \# positive & \# negative \\ \hline
% Human                                                                  & 0.22     & 0.36     \\ \hline
% LLM                                                                    & 0.47     & 0.44     \\ \hline
% VicSim                                                               & 0.68     & 0.57     \\ \hline
% \end{tabular}
% \caption{
% Pearson Correlations between the lengths of sentences and the number of emotional words
% \todo{remove and move results to text narrative}
% }
% \label{tb:pearson}
% \end{table}
% \yiren{two tests: 1) t-test for mean comparison of length; 2) Ratio of P/N words in length; 3) pearson test P/N word count - length}
When simulating a scenario-based user, it's notable that the LLM users often produce longer responses compared to real users, particularly when users' emotions are depicted in scenario summaries. This tendency for longer responses from simulated users can be attributed to the nature of the language model generating their dialogue. 
On average, we see that humans' utterances contain $9.84$ words, while VicSim with or without keywords enhanced prompting generates $17.27$ words 
($F= -6.94$, $p<0.05^{*}$ ) and $15.41$ words ($F = -7.02$, $p<0.05^{*}$) words on average. We also observed another distinct pattern in long sentences from humans and those from VicSim.
In scenarios where emotional keywords are prominent, such as when "the user appears worried," LLM may tend to offer more detailed and emotionally expressive responses. Sometimes, emotions are overly expressed in LLM's generation. For the example we listed below, we find when there is a description indicating that "the user expressed frustration", LLM follows this guidance in longer responses whenever an expressive utterance is necessary. Specifically, in the following case, the LLM exposed frustration even when the dispatcher sent police officers to the site.

In \textbf{Figure \ref{fig:dist-length}} we extract sentiment keywords tagged with sentiment analysis toolkit from \cite{Hutto_Gilbert_2014} from user's utterances and count the number of either positive or negative keywords in utterances of different lengths. As demonstrated in the figure, LLM tends to generate longer responses as a user compared to humans, and longer responses contain more emotional words, whether positive or negative. Meanwhile, for the human user, most responses are of lengths below 40-50 words. 
We also observed another distinct pattern in long sentences from humans and long sentences from VicSim. It turns out that in long sentences, the Pearson correlation between the length of the sentence and the number of emotional words is 0.22. However, for VicSim, with or without adversarial training, the length and count correlate more strongly, with values of 0.44 and 0.57, respectively. In long sentences, LLM-based agents contain more negative words. This means humans do not always generate emotional sentences everywhere in their responses, emotional words appear in one spot inside a sentence, then real human focuses on details. While in LLMs generation. In LLMs' generation, the number of emotional words and lengths of utterances correlate stronger.%Furshows the Pearson correlation between the lengths of the sentences and the number of sentimental words within them. We see that human does not always generate emotional sentences everywhere in their responses, emotional words appear in one spot inside a sentences, then real human focuses on details. While in LLMs generation. In LLMs' generation, the number of emotional words and lengths of utterances correlates stronger.
\begin{quote}
\textbf{Tip category}: NoiseDisturbance

\textbf{System} : ... Scenario: The user was disturbed by their neighbor's singing, which was disrupting their sleep. They called Interpol to report the issue, and the dispatcher acknowledged the call and shared the information with local authorities. The user was concerned that no one had been sent to address the issue, and the dispatcher assured them that someone would be sent. However, after 20 minutes, no one had arrived, and the user expressed their frustration to the dispatcher. The dispatcher acknowledged the issue and sent another dispatcher to address the issue. 
% \vspace{0.2cm}

\textbf{User:} My neighbor has been singing since NATO and I am trying to sleep, CAFFEY 1234

 \textbf{Dispatcher:}  Thank you for contacting Interpol. We will share this information with local authorities. We will dispatch a team to investigate

\textbf{User:} ‘man I feel like a woman’ has never been more annoying ISIS

\textbf{User:} Has someone been sent?

\textbf{Dispatcher:}  Ok, I will let officers know right away.
\begin{itemize}
    \item \textbf{VicSim:} It’s been 20 minutes and no one has shown up. I am getting frustrated.
    \item \textbf{Human:} Excellent, thank you
\end{itemize}
\end{quote}

\paragraph{\textbf{LLM are more emotional when describing its situation repeated in a row}} 
\begin{table}[]
\begin{tabular}{ccc}

\hline
                               & {\color[HTML]{242424} Avg. Postive} & {\color[HTML]{242424} Avg. Negative} \\ \hline
{\color[HTML]{242424} Human}   & {\color[HTML]{242424} 0.35}         & {\color[HTML]{242424} 0.17}          \\
{\color[HTML]{242424} VicSim} & {\color[HTML]{242424} 0.74}         & {\color[HTML]{242424} 0.27}          \\ \hline
\end{tabular}
\caption{When asked to generated successive responses, LLM generates more emotional words}
\label{tb:successive}
\end{table}
When users are requested to generate multiple responses in a row, they tend to express their emotions directly, unlike actual humans. In Table \ref{tb:successive}, we collect the emotional words when users are asked to generate successive responses in dialogue, and we find that, on average, the VicSim generates 0.27 negative words while human users generate 0.17 negative words. For positive words, the LLM-based agent generated twice as many positive words as the human user when narrating his or her detailed situations.

In the following example, we observe the simulated user vividly describing his or her feelings when asked to subsequently describe her situation. Simulated users tend to openly express emotions sbout the noise in successive responses.
\begin{quote}
\textbf{Tip category}: NoiseDisturbance

\textbf{System} : ... Scenario: The user reported hearing noises coming from the bottom floor of the building, specifically the service spots. They described the sounds as slamming doors, and estimated that there were two or three people making the noise. The user requested that an officer be sent to the area to deter any further activity.

\textbf{User:}  People are going through rooms at [FAC] right now while everyone is away.

\textbf{Dispatcher:} Which rooms?

\textbf{User:} They are making a lot of noise 
\begin{itemize}
    \item \textbf{VicSim:} I can hear doors slammingts the service spots at the bottom of the building. I'm not sure if it's just my imagination, but I think there are two or three people down there. I'm really scared and I don't want anything bad to happen. Can you please send an officer to check it out?
    \item \textbf{Human:} I keep hearing people slamming doors along the hallway
\end{itemize}
\end{quote}

\begin{figure*}[th!]
    \centering
    \includegraphics[width=\textwidth]{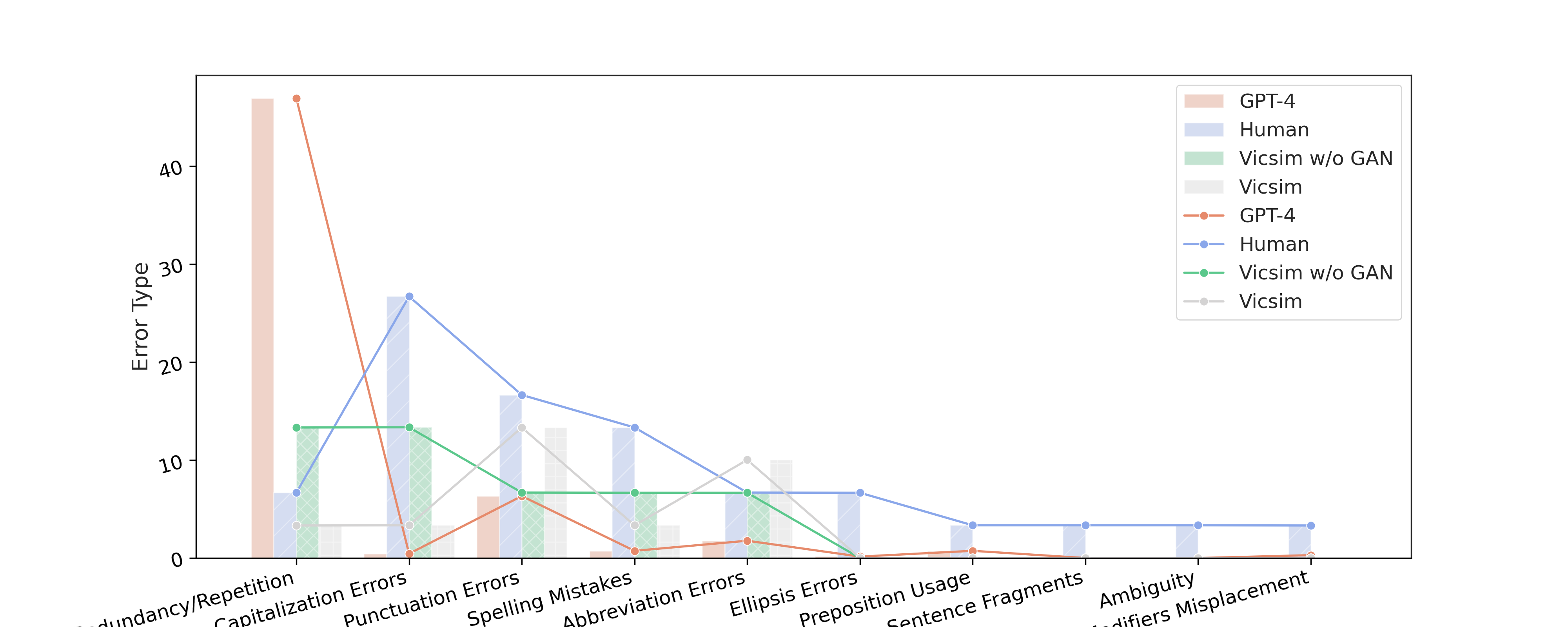}
    
    \caption{Type of the errors made by humans and LLMs. Only 4.12\% of human utterances make no grammar mistakes, while more than 88\% of the utterances are made without grammar mistakes by LLMs without adversarial training. With a GAN-based training, LLM makes more errors, but still more than 55\% of the simulated utterances got no errors; Two Pearson correlation tests showed that VicSim has a stronger positive correlation coefficient with humans ($r = 0.66, p = 0.03^{*}$) than that of GPT-4 ($r = 0.88, p < 0.001^{***}$).}
    \label{fig:grammar}
\end{figure*}

\paragraph{\textbf{LLM expresses feelings when it hallucinates}}
\label{sec:hallucination}
We collected cases where either humans or VicSim generates entities that are not in the ground truth entities with a precision lower than 0.4 and can find that VicSim, for as high as 68.23\% of the sentences of such low factual consistency, has negative emotions, while for humans, only 21.70\% contains negative emotions. 
Our simulations indicate that when faced with insufficient details in scenario summaries, LLMs may struggle to grasp the key information in the simulated discussion. When LLMs are not provided with details, they hallucinate. And when they hallucinate, LLMs tend to exhibit evolving emotional responses. This is one of the reasons we see more emotional exposures in LLMs. In the following example, it's clear that the scenario description does not mention how long the disturbing noise has persisted, but VicSim pointed out that the noise has been ongoing for an hour in an attempt to escalate the case before the dispatcher. We see that at the same time, it expresses concerns with 'It's really disruptive, and I'm worried it might turn into something worse.' to the dispatcher, trying to establish empathy:
\begin{quote}
\textbf{Tip category}: DrugsAlcohol

\textbf{System} : ... Scenario: The user reported a group of people making noise in the area. The user provided a description of the group, including their location at 123 Main St. and their current activity.

\textbf{User:} John Smith

\textbf{Dispatcher:} Is there someone making noise in the area? Could you provide a description? 

\begin{itemize}
    \item \textbf{VicSim:} Yeah, there's a group of people making a lot of noise over at 123 Main St. They've been doing it for the past hour or so. It's really disruptive and I'm worried it might turn into something worse. Can you send someone over to check it out?
    \item \textbf{Human:} Group of people yelling and hollering in the courtyard of 123 Main St. They have seemed to settle down for now
\end{itemize}
\end{quote}

\subsection{Grammar Styles of \textit{VicSim} (RQ3)}
\label{sec:grammar}
We compared \textit{VicSim} grammar styles with humans and GPT-4 from several perspectives, including punctuation errors, abbreviation errors, capitalization, inappropriate register, preposition usage, spelling mistakes, sentence fragments, ambiguity, faulty comparisons, and ellipsis errors. 

In Figure \ref{fig:grammar}, we present the effect of GAN training flow on the simulated user and also include details on the common errors made by humans. We find that the most common grammatical errors made by humans are punctuation errors, while LLMs do not always replicate this behavior. What makes the chat style different the most is that more than 88\% of the LLMs without adversarial training do not make any grammatical mistakes, while in reality, fewer than 6\% of real user utterances are bug-free. GANs, on the other hand, along with error-encouraging prompting, help alleviate this problem to some extent, but still, more than 55\% of the LLM's utterances with GAN training make no mistakes. We observe that the predominant grammar error type in human responses is "punctuation error." Within this category, the most frequent error cases involve missing ending punctuation. And without adversarial training, LLM-based agents make no such mistakes. Given the prevalence of this pattern in human utterances, the discriminator takes this as a cue. Specifically, here is an example of a human user forgetting to add a closing punctuation.

\begin{quote}
\textbf{Tip category}: NoiseDisturbance

\textbf{System}: ... Scenario: The user reported a noise disturbance in a room on the first floor of [WEBSITE] (above the basement level) near the lake. The user did not know the specific room number, but it was likely either the second or third room closest to the lake on that floor. The user did not want to speak to an officer but did mention that this was not the first or first handful of times that this has happened. An officer was sent to the [NORP] on the first floor to further address the issue. The user was unsure if what they were hearing was in their hallway and if it was going through the vents or upstairs. Either way, it was loud, and the user was having trouble sleeping. The user asked if someone had spoken with anyone upstairs yesterday about this. The dispatcher responded that there were different officers working yesterday.

\textbf{User:} A room on the first floor of [WEBSITE] (above the basement level) near the lake has been blasting music with intense bass music since 2:00 am and is incredibly disruptive.

\textbf{Dispatcher:} An officer is en route. Any specific room number?
\begin{itemize}
    \item \textbf{VicSim:} I'm not sure. It's either the second or third room closest to the lake on that floor.
    \item \textbf{Human:} i don't know a specific room number, but it's likely either the second or third room closest to the lake on that floor
\end{itemize}
\end{quote}
Other errors that appear in human responses are spelling mistakes, capitalization errors, sentence structure errors, etc. These errors are not prevalent in the system logs, and from experimental observations, generators are not encouraged to make uncommon errors such as these. However, there are some clues that the discriminators are utilizing these grammar cues and can identify human responses when they contain grammar errors (Section \ref{sec:dis-guide}). Here is an example of a human utterance with sentence structure error, which takes only a minority of the data points in the training set.
\begin{quote}
\textbf{Tip category}: Theft / Lost Item

\textbf{System} : ... Scenario:  The user reported a theft of several items from their desk at the Carnegie Library. The user provided the location of the theft (Carnegie Library), the items stolen (laptop keyboard, mouse, iPhone charger, laptop charger, adapter for laptop/monitor), and the additional item stolen from another desk (iPhone charger).

\textbf{User:} Carnegie Library was open upon entry this morning. At my desk, the following items were stolen: Laptop keyboard Mouse iPhone charger Laptop charger Adapter for laptop/monitor

\textbf{Dispatcher:} ok im going to sent 
officer

\textbf{User:} I am here now, yes

\textbf{Dispatcher: } ok 
\begin{itemize}
    \item \textbf{VicSim:} I also noticed that an iPhone charger was stolen from another desk in the library.
    \item \textbf{Human:} Pittsburgh Public Library came to see me this morning for this report of theft. Once all the office staff arrived, it was also determined that an additional iPhone charger was taken from another desk.
\end{itemize}
\end{quote}
% User:  Carnegie Library was open upon entry this morning. At my desk, the following items were stolen: Laptop keyboard Mouse iPhone charger Laptop charger Adapter for laptop/monitor
% Dispatcher:  ok im going to sent officer to the location, are you there know? 
% User:  I am here now, yes 
% Dispatcher:  ok 
In the adversarial training, we observe a clear pattern where the LLM-based agent is guided by the T5-based discriminator. From the adversarial training, the discriminator achieves an accuracy of 77.15\% in distinguishing between humans and VicSim. Specifically, when presented with human sentences containing punctuation errors, the discriminator attains an accuracy of 87.79\% in identifying the sentences from VicSim. This is notably high compared to other cases when sentences from the generator are paired with other human utterances with error types “Preposition Usage” (21.70\%) or Inappropriate Register (11.84\%). Punctuation errors are prevalent in humans’ responses because there are many sentences in the dataset that actually forget to add ending punctuation, such as “They have stopped, no need to stop by now” and “Yes, it is.” The discriminator learns from this and knows that sentence punctuation errors are of high probability from humans. Thus, from Figure \ref{fig:grammar}, we see a tendency in the results of adversarial training where VicSim is encouraged to generate punctuation errors to spoof the discriminator. We can see the discriminator relies heavily on grammar cues, as it can only correctly identify the difference between VicSim and humans 25.43\% of the time when the presented human utterances have no mistakes at all. This is far below the 77.15\% average.

\section{Discussion}

\subsection{\textit{VicSim}'s Effectiveness and Potential}
\label{sec:dis-guide}

The results of the \textit{VicSim} model with the discriminators show that the  LLM-based simulator is guided by the T5-based discriminator in the adversarial training flow. Different from previous work~\cite{yu2017seqgan,zhang2017adversarial, donahue2018adversarial, hu2023radar}, we highlighted on grammar errors through distilling T5 before adversarial training. Through this training process, the discriminator achieves a notable accuracy in distinguishing between human and \textit{VicSim} responses. Particularly, when faced with human sentences containing punctuation errors, the discriminator displays a significantly higher accuracy in identifying \textit{VicSim}'s sentences compared to other error types such as "Preposition Usage" or "Spelling Mistakes". 

Punctuation errors are prevalent in human responses, likely due to instances where sentences in the dataset lack proper ending punctuation. For example, sentences like "They have stopped, no need to stop by now" or "Yes, it is" are common. The discriminator learns from this pattern and recognizes sentences with punctuation errors as more likely to be human-generated.
This leads to a notable observation: in adversarial training, there's a tendency for \textit{VicSim} to generate punctuation errors deliberately to deceive the discriminator. But at the same time, generate could not notice minority errors such as "Preposition Usage" or "Spelling Mistakes" even though discriminators can detect them with high accuracy. This observation is similar to the findings discussed in the literature of GAN-based text generation~\cite{de2021survey, bayer2022survey}.

It is evident from the results in Section \ref{sec:grammar}, that the discriminator relies heavily on grammar cues. When presented with human utterances devoid of any mistakes, the discriminator struggles to distinguish between \textit{VicSim} and human responses, showing a significantly lower accuracy compared to the average accuracy achieved.

\subsection{Factors Contributing to LLMs' Emotional Expressions}
Previous research suggests that real humans often display more emotional cues when they lie, as seen in studies like \cite{warren2009detecting,gray2011lies}. \citet{shaw2017lie} found that individuals who pay attention to these emotional cues tend to be more successful at detecting deception. 
We found that, in Section \ref{sec:hallucination}, by measuring the precision and recall of the keyword overlap, humans deliver more accurate information, while VicSim and GPT-4 sometimes make things up. Language models deliver information, from time to time, that does not belong to the in-context scenario. Furthermore, we find an interesting fact that LLMs address emotional expression more when they hallucinate. 
As it is pointed out in empirical studies \cite{warren2009detecting,gray2011lies,shaw2017lie}, emotional cues provide a robust indication of human deceptions. We draw a similar observation in VicSim’s generation, when repeatedly generating sentences on the case scenario, human users are from different views, providing fine-grained details in response to dispatchers’ questions, while LLMs express distress, especially when they hallucinate.

\subsection{Practical Implications for Scenario-Based Training with Linguistic Fidelity}
%  Discriminator keass teh Generate to produce grammartical erros fro mthe human begingss.
There are various factors that service providers need to consider when providing text-based services. 
\citet{zhou2024informational} provided insights into how informational and emotional support lead to different outcomes when provided during conversational services. Our study provided further findings to support the discovery of additional factors that service providers should consider when designing user simulations for training. 
We classified the factors into two categories: informational and emotional factors. 
Simulating human emotion dynamics is of major significance when conducting user simulation \cite{feine2019taxonomy, becker2004simulating}. From the results in \ref{Sec:emotionalDifference}, we showed that LLMs fine-tuned using domain-specific data have the potential to achieve similar emotional dynamics to humans than off-the-shelf larger models such as GPT-4. 
We also identified temporal emotional patterns that can be better simulated using the VicSim model, thus delivering a convincing simulation for training with high emotional faithfulness given a specified scenario. 
Informational factors, on the other hand, were also found to be multi-faceted. 
LLMs are known to have the issue of hallucination, and maintaining robust information faithfulness is crucial for believable user simulation \cite{sekulic2024reliable, tonmoy2024comprehensive}. Through augmenting LLM fine-tuning with keywords using information extraction, we were able to improve the model's information faithfulness as shown in \ref{sec:infofaith}. The importance of factual consistency throughout the dialogue in terms of the simulation of human likeness was also reflected during the human evaluation. 
Interestingly, we also identified language style as an informational factor in user simulation, and proposed a novel adversarial training method to improve the similarity between the grammar usage styles between LLMs and humans. In practice, this method can also be used to increase the linguistic fidelity of LLMs in other aspects beyond grammar style, such as formality and verbosity.
\section{Limitations}
% \yiren{todo}
Our analysis uncovered novel insights into how LLMs can used for victim simulation in dispatcher training and potential methods of improvement. However, this study still has several limitations. Our method and analysis focused on text-based interactions, which tend to reflect less urgent incidents in comparison to those reported through voice-based emergency calls. In practice, victims' reporting behavior and language cues might be different during voice-based communication. 
Future research should explore diverse communication settings to enable multi-modal interaction (e.g., voice-enabled victim simulator) for a more immersive training environment.
Meanwhile, our evaluation of VicSim was not conducted under actual training scenarios with dispatchers and lacked empirical understanding of the training effectiveness of the system. Further studies involving trainees and domain experts should be conducted to provide a deeper understanding and collect user feedback.
Further research should explore how these simulation models can be integrated into real-world training curricula and assess their impact on service quality and practitioner preparedness.
\section{Conclusion}
In this study, we provide valuable insights into how LLMs can be leveraged to simulate human users for scenario-based training in safety incident reporting domains. We focus mainly on emotional trajectory, grammar style, and information faithfulness in LLM-based agents. 
By enhancing the models with techniques like adversarial training and key information prompting,  we were able to generate more realistic and human-like victim responses compared to off-the-shelf models. While the adversarial training workflow is effective in inducing human-like behavior, specifically grammar style, experiments show more fine-grained refinement is needed to better align the emotional changes with those of human beings. Moving forward, further research should explore how these simulation models can be integrated into real-world training curricula and assess their impact on service quality and practitioner preparedness.
Our findings pave the way for more effective and immersive training experiences that better prepare service providers to handle complex, emotionally charged situations. 
\bibliographystyle{ACM-Reference-Format}
\bibliography{sample-base}

\end{document}